\documentclass[10pt,conference]{IEEEtran}
\usepackage{cite}
\usepackage{booktabs}
\usepackage{multirow}
\usepackage{graphicx}
\usepackage{listings}
\usepackage{xcolor}
\usepackage{array}

\AtBeginDocument{%
	\setlength{\abovedisplayskip}{1pt plus 1pt minus 1pt}%
	\setlength{\belowdisplayskip}{1pt plus 1pt minus 1pt}%
	\setlength{\abovedisplayshortskip}{0pt plus 1pt minus 0pt}%
	\setlength{\belowdisplayshortskip}{0pt plus 1pt minus 0pt}%
}

\newcommand{\finding}[1]{\par\addvspace{0.15em}\noindent{\setlength{\fboxsep}{2pt}\fbox{\parbox{0.96\linewidth}{#1}}}\par\addvspace{0.15em}}
\let\oldthebibliography\thebibliography
\let\endoldthebibliography\endthebibliography
\renewenvironment{thebibliography}[1]{%
	\oldthebibliography{#1}%
	\fontsize{7.5}{8.2}\selectfont%
	\setlength{\itemsep}{0pt}%
	\setlength{\parsep}{0pt}%
	\setlength{\parskip}{0pt}%
}{\endoldthebibliography}
\newcolumntype{L}[1]{>{\raggedright\arraybackslash}p{#1}}
\newcolumntype{R}[1]{>{\raggedleft\arraybackslash}p{#1}}
\newif\ifshowlessons
\showlessonsfalse

\newcommand{\paperTitle}{Agent Skills Can Be Harmful: An Empirical Study of Skill-Induced Failures in LLM Agents}
\newcommand{\skillInducedFailure}{skill-induced agent failure}
\newcommand{\skillInducedFailures}{skill-induced agent failures}

\newcommand{\SkillInducedFailures}{Skill-induced agent failures}

\newcommand{\skillInducedFailureClasses}{skill-induced failures}

\newcommand{\corrCatApplicabilityFull}{Applicability Mismatch}
\newcommand{\corrCatApplicabilityAbbr}{APM}
\newcommand{\corrCatApplicabilityFirst}{\corrCatApplicabilityFull{} (\corrCatApplicabilityAbbr{})}

\newcommand{\corrCatEnvironmentFull}{Environment Mismatch}
\newcommand{\corrCatEnvironmentAbbr}{EM}
\newcommand{\corrCatEnvironmentFirst}{\corrCatEnvironmentFull{} (\corrCatEnvironmentAbbr{})}

\newcommand{\corrCatImplementationFull}{Task-Implementation Fault}
\newcommand{\corrCatImplementationAbbr}{TIF}
\newcommand{\corrCatImplementationFirst}{\corrCatImplementationFull{} (\corrCatImplementationAbbr{})}

\newcommand{\corrCatArtifactFull}{Artifact Misplacement}
\newcommand{\corrCatArtifactAbbr}{AM}
\newcommand{\corrCatArtifactFirst}{\corrCatArtifactFull{} (\corrCatArtifactAbbr{})}

\newcommand{\corrRootBrokenEnvFull}{Broken Dependency or Runtime}
\newcommand{\corrRootBrokenEnvAbbr}{BDR}
\newcommand{\corrRootBrokenEnvFirst}{\corrRootBrokenEnvFull{} (\corrRootBrokenEnvAbbr{})}

\newcommand{\corrRootIncompatEnvFull}{Environment-State Mismatch}
\newcommand{\corrRootIncompatEnvAbbr}{ESM}
\newcommand{\corrRootIncompatEnvFirst}{\corrRootIncompatEnvFull{} (\corrRootIncompatEnvAbbr{})}

\newcommand{\corrRootWorkflowFull}{Obstructive Workflow Guidance}
\newcommand{\corrRootWorkflowAbbr}{OWG}
\newcommand{\corrRootWorkflowFirst}{\corrRootWorkflowFull{} (\corrRootWorkflowAbbr{})}

\newcommand{\corrRootApproachFull}{Incorrect Required-Element Fill}
\newcommand{\corrRootApproachAbbr}{IRF}
\newcommand{\corrRootApproachFirst}{\corrRootApproachFull{} (\corrRootApproachAbbr{})}

\newcommand{\corrRootOmissionFull}{Required-Element Omission}
\newcommand{\corrRootOmissionAbbr}{RRO}
\newcommand{\corrRootOmissionFirst}{\corrRootOmissionFull{} (\corrRootOmissionAbbr{})}

\newcommand{\corrRootLocationFull}{Wrong Artifact Location}
\newcommand{\corrRootLocationAbbr}{WAL}
\newcommand{\corrRootLocationFirst}{\corrRootLocationFull{} (\corrRootLocationAbbr{})}

\newcommand{\perfCatContextFull}{Context Bloat}
\newcommand{\perfCatContextAbbr}{CO}
\newcommand{\perfCatContextFirst}{\perfCatContextFull{} (\perfCatContextAbbr{})}

\newcommand{\perfCatProcedureFull}{Excessive Procedure}
\newcommand{\perfCatProcedureAbbr}{EP}
\newcommand{\perfCatProcedureFirst}{\perfCatProcedureFull{} (\perfCatProcedureAbbr{})}

\newcommand{\perfCatDependencyFull}{Dependency Resolution}
\newcommand{\perfCatDependencyAbbr}{DO}
\newcommand{\perfCatDependencyFirst}{\perfCatDependencyFull{} (\perfCatDependencyAbbr{})}

\newcommand{\perfRootBodyBloatFull}{Skill-Body Context Bloat}
\newcommand{\perfRootBodyBloatAbbr}{SBCB}
\newcommand{\perfRootBodyBloatFirst}{\perfRootBodyBloatFull{} (\perfRootBodyBloatAbbr{})}

\newcommand{\perfRootSuppBloatFull}{Supplementary-Material Bloat}
\newcommand{\perfRootSuppBloatAbbr}{SMB}
\newcommand{\perfRootSuppBloatFirst}{\perfRootSuppBloatFull{} (\perfRootSuppBloatAbbr{})}

\newcommand{\perfRootExplorationFull}{Excessive Exploration}
\newcommand{\perfRootExplorationAbbr}{EE}
\newcommand{\perfRootExplorationFirst}{\perfRootExplorationFull{} (\perfRootExplorationAbbr{})}

\newcommand{\perfRootPipelineFull}{Heavy Implementation Pipeline}
\newcommand{\perfRootPipelineAbbr}{HIP}
\newcommand{\perfRootPipelineFirst}{\perfRootPipelineFull{} (\perfRootPipelineAbbr{})}

\newcommand{\perfRootVerificationFull}{Excessive Verification}
\newcommand{\perfRootVerificationAbbr}{EV}
\newcommand{\perfRootVerificationFirst}{\perfRootVerificationFull{} (\perfRootVerificationAbbr{})}

\newcommand{\funcFail}{functional failure}
\newcommand{\funcFails}{functional failures}
\newcommand{\FuncFail}{Functional failure}

\newcommand{\effReg}{efficiency regression}
\newcommand{\effRegs}{efficiency regressions}
\newcommand{\EffReg}{Efficiency regression}

\newcommand{\funcFailMech}{What are the root causes of \funcFails{}?}
\newcommand{\effRegMech}{What are the root causes of \effRegs{}?}

\newcommand{\funcFailTax}{functional-failure taxonomy}

\newcommand{\allFailureCount}{307}
\newcommand{\corrTotalCount}{125}
\newcommand{\corrTotalPct}{100.0\%}
\newcommand{\perfTotalCount}{182}
\newcommand{\perfTotalPct}{100.0\%}
\newcommand{\corrBaselineCount}{38}

\newcommand{\corrCrossSkillCount}{87}

\newcommand{\perfBaselineCount}{128}

\newcommand{\perfCrossSkillCount}{54}

\newcommand{\corrImplementationCount}{86}
\newcommand{\corrImplementationPct}{68.8\%}
\newcommand{\corrEnvironmentCount}{13}
\newcommand{\corrEnvironmentPct}{10.4\%}
\newcommand{\corrApproachOmissionCount}{82}

\newcommand{\perfProcedureCount}{114}
\newcommand{\perfProcedurePct}{62.6\%}
\newcommand{\perfContextCount}{46}
\newcommand{\perfContextPct}{25.3\%}

\newcommand{\corrMetadataCount}{2}
\newcommand{\corrMetadataPct}{1.6\%}
\newcommand{\corrBrokenEnvCount}{5}
\newcommand{\corrBrokenEnvPct}{4.0\%}
\newcommand{\corrIncompatEnvCount}{8}
\newcommand{\corrIncompatEnvPct}{6.4\%}
\newcommand{\corrWorkflowCount}{4}
\newcommand{\corrWorkflowPct}{3.2\%}
\newcommand{\corrApproachCount}{46}
\newcommand{\corrApproachPct}{36.8\%}
\newcommand{\corrOmissionCount}{36}
\newcommand{\corrOmissionPct}{28.8\%}
\newcommand{\corrLocationCount}{24}
\newcommand{\corrLocationPct}{19.2\%}

\newcommand{\perfBodyBloatCount}{43}
\newcommand{\perfBodyBloatPct}{23.6\%}
\newcommand{\perfSuppBloatCount}{3}
\newcommand{\perfSuppBloatPct}{1.6\%}
\newcommand{\perfExplorationCount}{17}
\newcommand{\perfExplorationPct}{9.3\%}
\newcommand{\perfPipelineCount}{30}
\newcommand{\perfPipelinePct}{16.5\%}
\newcommand{\perfVerificationCount}{67}
\newcommand{\perfVerificationPct}{36.8\%}

\newcommand{\perfDependencyRepairCount}{22}
\newcommand{\perfDependencyRepairPct}{12.1\%}

\newcommand{\attribVoteBaselineRoot}{35/38}
\newcommand{\attribVoteBaselineRootPct}{92.1\%}
\newcommand{\attribVoteBaselineCategory}{37/38}
\newcommand{\attribVoteBaselineCategoryPct}{97.4\%}
\newcommand{\attribVoteBaselineCthree}{15/17}
\newcommand{\attribVoteBaselineCthreePct}{88.2\%}
\newcommand{\attribVoteCrossRoot}{76/87}
\newcommand{\attribVoteCrossRootPct}{87.4\%}
\newcommand{\attribVoteCrossCategory}{80/87}
\newcommand{\attribVoteCrossCategoryPct}{92.0\%}
\newcommand{\attribVoteCrossCthree}{61/69}
\newcommand{\attribVoteCrossCthreePct}{88.4\%}
\newcommand{\attribVoteCombinedRoot}{111/125}
\newcommand{\attribVoteCombinedRootPct}{88.8\%}
\newcommand{\attribVoteCombinedCategory}{117/125}
\newcommand{\attribVoteCombinedCategoryPct}{93.6\%}
\newcommand{\attribVoteCombinedCthree}{76/86}
\newcommand{\attribVoteCombinedCthreePct}{88.4\%}

\begin{document}

\title{\paperTitle{}}

\author{
\IEEEauthorblockN{Gen Dong$^{1}$,
Yanjie Gao$^{2,\dagger}$,
Liqun Li$^{3}$,
Tianyin Xu$^{4}$,
Yu Hua$^{1}$,
Fan Yang$^{2}$}
\IEEEauthorblockA{$^{1}$Huazhong University of Science and Technology}
\IEEEauthorblockA{$^{2}$Microsoft Research}
\IEEEauthorblockA{$^{3}$Microsoft}
\IEEEauthorblockA{$^{4}$University of Illinois Urbana-Champaign}}

\maketitle
\begingroup
\renewcommand{\thefootnote}{\fnsymbol{footnote}}
\footnotetext[0]{This work was done during Gen Dong's internship at Microsoft Research. $^{\dagger}$Corresponding author.}
\endgroup

\begin{abstract}
	Agent skills are the {\it de facto} mechanism for extending LLM agents with reusable guidance.
	A skill can shape the agent's task execution, including planning, tool use, problem-solving,
	and validation.
	Prior work reported mixed results of agent skills: some skills improve task success rates,
	while others have no effect, increase token use and execution time, and even reduce success rates.
	This paper presents a comprehensive analysis of \skillInducedFailures{}
	by attributing task failures and cost regressions to specific loaded skills.
	We introduce a differential analysis framework that attributes a failure or regression to a skill by comparing a target skill-guided run against a no-skill or semantically matched-skill reference run that solves the same task, or solves it more cheaply.
	We instantiate this framework on SkillsBench and SWE-Skills-Bench, yielding \allFailureCount{} \skillInducedFailureClasses{},
	including \corrTotalCount{} \funcFails{} and \perfTotalCount{} \effRegs{}.
	We also build \textsc{SkillTriage}, a taxonomy-guided attribution tool that normalizes paired cases, extracts differential evidence, and produces triage reports.
	Our major findings include:
	(1) Skill-induced \funcFails{} are rarely caused by obviously irrelevant skills; instead, seemingly relevant skills often make the agent incorrectly implement or omit task-required implementation elements.
	\corrCatImplementationFull{} accounts for \corrImplementationCount{} of \corrTotalCount{} \funcFails{} (\corrImplementationPct{}), while wrong artifact locations and environment mismatches account for \corrLocationCount{} and \corrEnvironmentCount{} cases, respectively.
	(2) Skill-induced \effRegs{} are not explained by prompt length alone.
	When regressions come from context overhead, they are almost entirely caused by mandatory skill-body text (\perfBodyBloatCount{} of \perfContextCount{} context-overhead cases), but overall \perfCatProcedureFull{} dominates with \perfProcedureCount{} of \perfTotalCount{} cases (\perfProcedurePct{}).
	(3) The largest sources within \perfCatProcedureFull{} are excessive verification and heavy implementation pipelines, contributing \perfVerificationCount{} and \perfPipelineCount{} cases, respectively.
	This shows that skills often turn validation checklists and construction recipes into mandatory work.
	Based on our findings, we propose research topics and tooling improvements for safer and more cost-aware skill reuse.
\end{abstract}

% \begin{IEEEkeywords}
% LLM agents, agent skills, empirical study, software engineering, performance regressions, context engineering
% \end{IEEEkeywords}

\section{Introduction}
\label{sec:intro}
LLM agents are increasingly used to solve complex software-engineering and other knowledge-intensive tasks by combining model reasoning with tools, repositories, and external environments.
To make procedural guidance reusable across tasks, recent systems introduce \emph{skills}: instruction packages that encode task-specific procedures, examples, constraints, and tool-use guidance.
Once selected, a skill is inserted into the agent context and can shape planning, file edits, command execution, validation, and stopping decisions.

However, this mechanism also creates a risk.
A skill may be topically relevant but operationally wrong for the concrete task: it may imply an incompatible path, over-prescribe a workflow, induce an incompatible dependency or runtime state, or encourage unnecessary verification.
Such guidance can induce a \funcFail{} when the task fails, or an \effReg{} when the task passes only with substantially higher token use or longer execution time.
These failures are especially problematic because skills are intended to be reused, so a harmful skill can repeatedly bias future agent executions.

Recent skill benchmarks suggest that this risk is not hypothetical.
SkillsBench reports average gains from curated skills but also observes pass-rate drops on 16 of 84 tasks~\cite{li2026skillsbench}.
SWE-Skills-Bench reports limited average improvement, many skills with no pass-rate gain, and cases with token overhead up to 451\%~\cite{han2026sweskillsbench}.
These results show that skills can help, harm, or waste resources depending on the task and skill content.

However, existing evaluations leave four gaps that make skill-induced failures difficult to study.
First, the central challenge is not merely detecting failed runs, but isolating and attributing failures to the loaded skill. A failed with-skill run alone cannot distinguish skill-induced harm from base-agent limitations, verifier scope, ordinary agent variance, or environmental artifacts. Identifying skill-induced failures therefore requires contrastive evidence.
Second, existing skill benchmarks are designed to measure utility rather than uncover failure mechanisms. Consequently, their original skill settings provide limited negative cases for studying skill-induced failures.
Third, aggregate pass rates and cost ratios provide outcome-level metrics but do not explain how a skill changes an agent trajectory. Related studies on long context, irrelevant context, and prompt sensitivity explain broad context effects~\cite{liu2024lostmiddle,shi2023distracted,sclar2024sensitivity,levy2025moredocs}. However, they do not identify how a specific skill document causes a verifier-facing \funcFail{} or a hidden \effReg{}.
Finally, even when failure attribution is performed manually, it does not scale to agent platforms or skill marketplaces where new skills and skill updates must be screened continuously.

In this paper, we present a comprehensive empirical study of \skillInducedFailures{} in LLM agents.
Inspired by differential testing~\cite{mckeeman1998differential}, we construct paired target/reference executions that use a no-skill or semantically matched skill execution as a pseudo-oracle.
The design attributes a failure to a skill only when the same task is otherwise solvable or can be solved more cheaply under this paired-comparison setup.
We instantiate the study on SkillsBench and SWE-Skills-Bench, augment the original benchmark settings with semantically matched public skills, and analyze task specifications, skill contents, execution trajectories, verifier outcomes, token use, and execution time.
We then derive root-cause taxonomies for \funcFails{} and \effRegs{}, and build \textsc{SkillTriage}, a taxonomy-guided tool that normalizes paired cases, extracts differential evidence, and produces attribution reports for triage.

Specifically, our study addresses the following research questions:
\begin{enumerate}
	\item \textbf{RQ1: How can we construct a contrastive dataset of \skillInducedFailures{} for root-cause analysis?}
	      To answer this question, we design a differential-testing-motivated contrastive construction that pairs each target skill-guided execution with a no-skill or semantically matched-skill reference run as a pseudo-oracle.
	      Using our contrastive labeling criteria, we confirm \allFailureCount{} \skillInducedFailureClasses{}, including \corrTotalCount{} \funcFails{} and \perfTotalCount{} high-confidence \effRegs{}.
	      A detailed description is provided in Section~\ref{sec:method}.

	\item \textbf{RQ2: What are the root causes of skill-induced \funcFails{}?}
	      To answer this question, we manually analyze task specifications, skill contents, execution trajectories, verifier outputs, and artifacts for the confirmed \funcFails{}.
	      Our study reveals several notable findings.
	      For example, most \funcFails{} are not caused by obviously irrelevant skills.
	      Instead, seemingly relevant skills often make the agent implement a required field, API behavior, calculation, output format, or domain rule incorrectly or omit it entirely.
	      \corrCatImplementationFull{} accounts for \corrImplementationCount{} of \corrTotalCount{} \funcFails{} (\corrImplementationPct{}), while verifier-facing artifact and environment boundaries account for \corrLocationCount{} and \corrEnvironmentCount{} cases, respectively.
	      A detailed analysis is provided in Section~\ref{sec:rq2}.

	\item \textbf{RQ3: What are the root causes of skill-induced \effRegs{}?}
	      To answer this question, we analyze target/reference pairs whose token use and execution time both increase under the loaded skill, with at least one metric exceeding the primary $T=2.0$ threshold.
	      Our study reveals that high-confidence \effRegs{} are dominated by \perfCatProcedureFull{} rather than prompt length alone.
	      At the primary $T=2.0$ threshold, \perfCatProcedureFull{} accounts for \perfProcedureCount{} of \perfTotalCount{} cases (\perfProcedurePct{}), driven mainly by excessive verification (\perfVerificationCount{} cases) and heavy implementation pipelines (\perfPipelineCount{} cases).
	      A detailed analysis is provided in Section~\ref{sec:rq3}.

	\item \textbf{RQ4: How can we automatically attribute \skillInducedFailures{} to their root causes?}
	      To answer this question, we build \textsc{SkillTriage}, a taxonomy-guided attribution tool that normalizes paired cases, extracts differential evidence, and produces triage reports.
	      For \funcFails{} and \effRegs{}, it matches manually assigned exact root causes for \attribVoteCombinedRoot{} (\attribVoteCombinedRootPct{}) and 132/182 (72.5\%), respectively.
	      A detailed evaluation is provided in Section~\ref{sec:attribution}.
\end{enumerate}

In summary, this paper makes the following contributions:
\begin{enumerate}
	\item
	      We identify a critical and timely problem in LLM-agent skill reuse: skills can cause \funcFails{} or substantial \effRegs{} even when they are topically relevant.
	      We construct a contrastive analysis dataset of \allFailureCount{} confirmed \skillInducedFailureClasses{} from SkillsBench and SWE-Skills-Bench, where each case is validated by no-skill or semantically matched-skill reference runs and annotated as a \funcFail{} or \effReg{}.

	\item
	We present the first comprehensive study of skill-induced failures in LLM agents, categorizing \funcFails{} and \effRegs{} and analyzing their root causes.

	\item
	      We develop a tool, \textsc{SkillTriage}, which normalizes paired cases, extracts differential evidence, and attributes \skillInducedFailures{} to root causes.
	      We demonstrate the practical utility and effectiveness of \textsc{SkillTriage} through automated attribution evaluation.
\end{enumerate}

\section{Background}
\label{sec:background}

\paragraph{Agent skills in LLM agents}
An \emph{agent skill} is a structured package of procedural knowledge that augments an LLM agent at inference time without modifying the model parameters~\cite{anthropic2025agentskills}.
In current agent systems, a skill is usually organized around a named \texttt{SKILL.md} file.
The skill framework selects one or more skills for a task and loads their contents into the agent context.
Once loaded, a skill can influence the agent's planning, repository exploration, tool use, code edits, tests, and stopping decision.

\paragraph{Skill contents}
A skill typically contains four kinds of information.
First, metadata such as the skill name, description, frontmatter, or when-to-use text signals when the skill applies.
Second, the main body gives procedural guidance, including task-solving steps, implementation rules, workflow constraints, and checklists.
Third, supplementary files may provide examples, templates, schemas, scripts, reference documents, or other reusable resources loaded on demand.
Fourth, validation guidance may suggest tests, expected outputs, debugging steps, or completion criteria.
These components matter because they become part of the agent's effective task context rather than passive documentation.

\begin{figure}[t]
\begin{lstlisting}[language={},basicstyle=\ttfamily\scriptsize]
<!-- (*@\textbf{\texttt{Metadata:}}@*) -->
---
name: rag-backend-helper
description: Use for backend RAG applications.
---
<!-- (*@\textbf{\texttt{Skill name:}}@*) -->
# RAG Backend Helper
<!-- (*@\textbf{\texttt{Procedure body:}}@*) -->
1. Inspect the repository structure and identify the app package.
2. Install or import vector-store and document-loader packages.
3. Add retrieval code under the app's retriever module.
<!-- (*@\textbf{\texttt{Validation guidance:}}@*) -->
4. Run the full test suite and fix all failures before stopping.
<!-- (*@\textbf{\texttt{Supplementary material:}}@*) -->
See examples/rag_template.py for a reference implementation.
\end{lstlisting}
\caption{A simplified \texttt{SKILL.md} example showing typical skill components.}
\label{fig:skill-example}
\end{figure}

Figure~\ref{fig:skill-example} illustrates the kind of information that a skill can inject.
The description provides an applicability signal; repository and module wording can shape where the agent writes artifacts; dependency instructions can alter the execution environment; and verification checklists can add procedure cost.
Thus, a skill can help by providing reusable knowledge, but it can also mislead the agent when its assumptions do not match the concrete task.

\paragraph{Agent Execution Trajectories}
An \emph{execution trajectory} is the ordered sequence of model interactions, tool invocations, and intermediate actions produced by an agent while attempting a task.
A trajectory may include model responses, tool calls, file operations, searches, code edits, command executions, and other observable interactions with the environment.
Throughout this paper, root-cause analysis is performed based on execution trajectories rather than solely on final task outcomes.

\section{Methodology}
\label{sec:method}

\subsection{Study Design}
\begin{figure}[t]
	\centering
	\includegraphics[width=\linewidth]{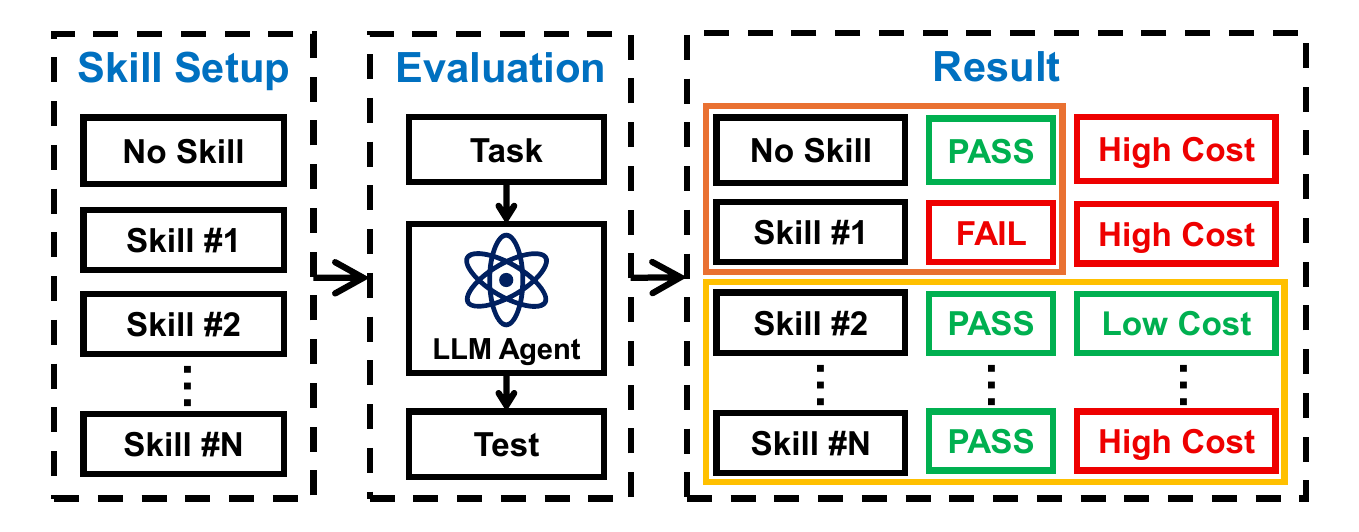}
	\caption{Differential evaluation across skill setups.}
	\label{fig:differential-evaluation}
\end{figure}

We use a differential-testing-inspired contrastive design to attribute failures to loaded skills rather than to baseline model limitations, task difficulty, or ordinary differences between repeated executions.
Figure~\ref{fig:differential-evaluation} illustrates the setup: paired executions keep the task, verifier, agent framework, model, repository or container state, and input data fixed, and vary only the skill setup, including no skill and candidate skills.
Each run produces two observable outcomes: a correctness outcome, measured by the task verifier or tests, and a cost outcome, measured by token use and execution time.

We define the run being audited as the \emph{target run}.
A \emph{reference run} is another execution of the same task under a different skill setup, either no skill or a semantically matched skill.
Throughout the paper, \textsc{PASS} and \textsc{FAIL} denote the deterministic verifier outcome of a single run. 
For a paired comparison, \textsc{FAIL}/\textsc{PASS} means that the target run fails and the reference run passes, while \textsc{PASS}/\textsc{PASS} means that both runs pass.
We use two audit settings throughout the paper. A \emph{with/no-skill comparison} compares a target run executed with the audited skill against a reference run on the same task without any skill. A \emph{cross-skill comparison} compares the target run against another semantically matched skill setup on the same task. For \funcFails{}, the target run fails while the reference run passes; for \effRegs{}, both runs pass but the target run is substantially more expensive.
The reference run acts as a pseudo-oracle, not as a ground-truth solution: it shows that the same task can be solved, or solved more cheaply, under the same task and verifier conditions.
With task, verifier, repository, model, and agent framework fixed, only the skill setup varies; therefore, when the reference run passes or solves the task more efficiently, it provides contrastive evidence for attributing the difference to the loaded skill under controlled conditions.

We define two failure classes:
\begin{itemize}
	\item \textbf{\FuncFail{}}: the target run fails the verifier while a reference run passes.
	\item \textbf{\EffReg{}}: both the target and reference runs pass, but the target run has substantially more token use, longer execution time, or both.
\end{itemize}
In Figure~\ref{fig:differential-evaluation}, the orange box highlights the \funcFail{} pattern: the no-skill run passes, but the audited skill run fails on the same task. The yellow box highlights the \effReg{} pattern: two runs both pass, but the audited skill run incurs substantially higher cost than the reference run.

\begin{figure*}[t]
	\centering
	\includegraphics[width=0.96\textwidth]{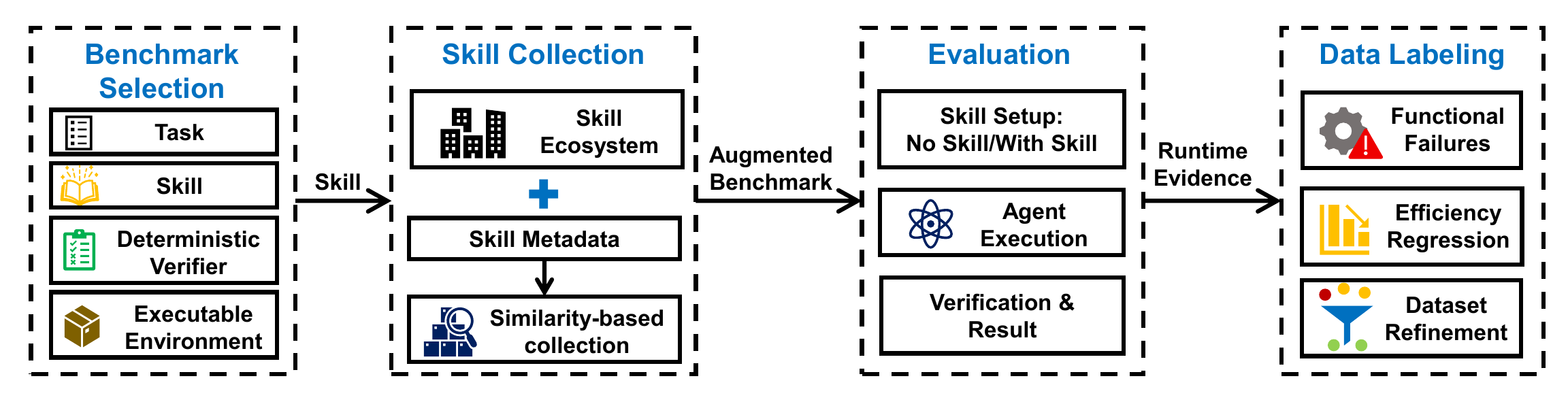}
	\caption{Overview of the study methodology.}
	\label{fig:method-overview}
\end{figure*}

Figure~\ref{fig:method-overview} summarizes our four-stage methodology.
First, we select skill benchmarks with deterministic verifiers and executable task environments, so that pass/fail and cost outcomes can be measured reproducibly rather than decided by manual judgment after the run.
Second, we augment these benchmarks with semantically matched public skills, broadening the skill-comparison space and making it possible to observe a wider range of skill-induced failures.
Third, we execute the augmented benchmark tasks with the agent under controlled skill setups and collect runtime evidence, including trajectories, verifier results, token use, and execution time.
Finally, we label \funcFails{} and \effRegs{} from the paired execution data and refine the dataset by removing ambiguous, verifier-narrow, or duplicate cases before root-cause analysis.

\subsection{Study Subjects}
Our study uses two representative skill benchmarks as study subjects.
SkillsBench provides 84 evaluated tasks across 11 domains with no-skill, curated-skill, and self-generated-skill settings~\cite{li2026skillsbench}.
SWE-Skills-Bench provides 490 repository-based software-engineering task instances with no-skill and curated-skill settings~\cite{han2026sweskillsbench}.
Both benchmarks use deterministic programmatic verifiers, enabling stable pass/fail and cost comparisons.

The original benchmark settings are valuable starting points because they provide executable tasks, deterministic verifiers, and curated skills, but they are not sufficient by themselves for mechanism-level failure attribution.
Their primary goal is to measure skill utility rather than to mine failure mechanisms, so the comparison space is narrow: SkillsBench compares no-skill, curated-skill, and self-generated-skill conditions, while SWE-Skills-Bench compares no-skill and curated-skill conditions.
These settings can show that a skill helps, hurts, or adds cost, but they often do not provide an alternative successful skill for the same task, which is needed to localize the failure to differences in skill content and induced trajectories.

\subsection{Data Collection}
To broaden the skill setups available for paired evaluation beyond the original benchmark settings, we search two public skill-sharing sites that host reusable agent skills: \texttt{smithery.ai} and \texttt{skillsmp.com}.
We use these sites because their skills include agent-visible names and descriptions, which allow us to retrieve plausible alternatives to the curated benchmark skills.
For each curated benchmark skill, we query the sites with the curated skill's name and description, embed the curated and candidate skill metadata with \texttt{all-MiniLM-L6-v2}, a widely used sentence-embedding model for semantic-similarity matching, and retain up to the top-5 candidates with cosine similarity of at least 0.7 to filter out off-topic matches while limiting evaluation cost.
We refer to these retained metadata-similar candidates as semantically matched public skills.

The retrieved skills are used as candidate skill setups, not as mandatory context loaded regardless of the task.
During execution, the agent receives the task and the available candidate skill package under the normal skill-loading interface; it may choose to load the skill when it judges the skill relevant, or avoid relying on it when the task does not match.
This design reduces the risk that an unrelated retrieved skill is artificially forced into the agent context, while still expanding the space of plausible skill alternatives for contrastive analysis.
For SkillsBench, the original self-generated-skill condition is included in the original comparison space but not treated as a variant in the expanded public-skill comparison space, because our expansion focuses on reusable public skills that users are more likely to select and load rather than benchmark-specific self-generated skills.

Together with the original curated skill, these retained public variants give up to six skill variants per task.
For a task with $k$ available skill variants, with/no-skill comparisons contribute $k$ no-skill/with-skill pairs, and cross-skill comparisons contribute $k(k-1)$ ordered skill/skill pairs because each skill variant can define the audited target run while another variant defines the reference run.
In the original benchmark settings, SkillsBench has two with-skill conditions per task, while SWE-Skills-Bench has one; after expansion, both benchmarks have up to six skill variants per task. Table~\ref{tab:comparison-space} summarizes the resulting comparison space before evaluation.
The augmentation therefore increases the comparison space from 826 to 20{,}664 potential paired comparisons, a roughly $25\times$ expansion, making mechanism-level failure attribution feasible.

\begin{table}[t]
	\caption{Potential paired comparisons before evaluation.}
	\label{tab:comparison-space}
	\centering\scriptsize
	\setlength{\tabcolsep}{2pt}
	\renewcommand{\arraystretch}{1.08}
	\begin{tabular}{|L{2.3cm}|L{1.45cm}|R{1.45cm}|R{1.65cm}|}
		\hline
		\textbf{Benchmark} & \textbf{Pair type} & \textbf{Original} & \textbf{Expanded} \\
		\hline
		SkillsBench        & With/no-skill           & 168               & 504               \\
		\hline
		SkillsBench        & Cross-skill        & 168               & 2{,}520           \\
		\hline
		SWE-Skills-Bench   & With/no-skill           & 490               & 2{,}940           \\
		\hline
		SWE-Skills-Bench   & Cross-skill        & 0                 & 14{,}700          \\
		\hline
\multicolumn{2}{|l|}{Total} & 826 & 20{,}664 \\

		\hline
	\end{tabular}
\end{table}

We execute tasks with OpenCode 1.15.1~\cite{opencode2026github} and Claude Opus 4.6~\cite{anthropic2026opus46} under with/no-skill and cross-skill comparison settings. This configuration combines a widely used open-source agent runtime with a state-of-the-art frontier language model, providing a representative setup for contemporary LLM agents.
Each pair fixes the task instruction, repository or container state, data files, verifier, agent framework, and model; only the skill setup changes.
For each run, we record the loaded skill, trajectory, verifier outcome, token use, and execution time.

\subsection{Data Labeling}
\paragraph{Functional failures}
We include two types of \funcFails{} in the final analysis dataset.
A with/no-skill-comparison \funcFail{} requires the no-skill run to pass and the audited-skill run to fail.
A cross-skill \funcFail{} requires the target run with the audited skill to fail and a run with another semantically matched skill to pass.

\paragraph{Efficiency regressions}
For \effRegs{}, we only compare \textsc{PASS}/\textsc{PASS} pairs, where both the target and reference runs pass.
Let $r_{\mathrm{tok}}$ be the target-to-reference token ratio and $r_{\mathrm{time}}$ be the target-to-reference execution-time ratio. We classify a \textsc{PASS}/\textsc{PASS} pair as an \effReg{} at threshold $T$ iff:
\[
\min(r_{\mathrm{tok}}, r_{\mathrm{time}}) > 1.0
\;\land\;
\max(r_{\mathrm{tok}}, r_{\mathrm{time}}) > T .
\]

That is, both token use and execution time must regress relative to the reference run, and at least one must more than double ($T = 2.0$) relative to the reference run.
Requiring both ratios to be greater than $1.0$ excludes token--time tradeoffs, where one metric increases while the other decreases.
We use $T=2.0$ as the primary threshold to focus on large cost regressions and reduce sensitivity to small token/time fluctuations. We label each \effReg{} as \emph{token-dominant}, \emph{time-dominant}, or \emph{joint token-and-time} according to which ratio exceeds $T$.

\paragraph{Dataset refinement}
After data collection, we identify \funcFails{} and \effRegs{} from the executed pairs in the expanded comparison space as labeled candidates. From the 20{,}664 potential paired comparisons summarized in Table~\ref{tab:comparison-space}, the executed evaluations yield 665 labeled candidates: 315 \funcFail{} candidates and 350 \effReg{} candidates.
We remove candidates with insufficient evidence, likely verifier-induced false positives, and duplicate same-task/same-skill effects. When the same failure appears in both with/no-skill and cross-skill audits, we keep the with/no-skill instance. 
After refinement, we manually inspect each remaining case, assign one root-cause label, and finalize the labels through group consensus.
Table~\ref{tab:failure-dataset-flow} summarizes the count flow from labeled candidates to final analysis cases.
For \funcFails{}, the with/no-skill audit starts from 70 automatically identified with-skill-\textsc{FAIL}/no-skill-\textsc{PASS} candidate records: 50 from SWE-Skills-Bench and 20 from SkillsBench.
The cross-skill audit starts from 245 failed with-skill records in tasks where at least one skill run passes and another skill run fails: 131 from SWE-Skills-Bench and 114 from SkillsBench.
For \effRegs{}, labeled candidates are threshold-triggering cost-regression candidates identified under the primary threshold $T=2.0$ before dataset refinement: 159 from with/no-skill audits and 191 from cross-skill audits.
\begin{table}[t]
	\caption{Failure dataset construction from labeled candidates to final analysis cases.}
	\label{tab:failure-dataset-flow}
	\centering\footnotesize
	\setlength{\tabcolsep}{2pt}
	\begin{tabular}{|L{2.1cm}|L{1.7cm}|R{1.7cm}|R{1.7cm}|}
		\hline
		\textbf{Failure Type} & \textbf{Audit setting} & \textbf{Labeled candidates} & \textbf{Analysis cases} \\
		\hline
		\multirow{3}{*}{\shortstack[l]{Functional                                                      \\failure}} & With/no-skill & 70 & \corrBaselineCount{} \\
		\cline{2-4}
		              & Cross-skill            & 245                         & \corrCrossSkillCount{}  \\
		\cline{2-4}
		              & Subtotal               & 315                         & \corrTotalCount{}       \\
		\hline
		\multirow{3}{*}{\shortstack[l]{Efficiency                                                      \\regression}} & With/no-skill & 159 & \perfBaselineCount{} \\
		\cline{2-4}
		              & Cross-skill            & 191                         & \perfCrossSkillCount{}  \\
		\cline{2-4}
		              & Subtotal               & 350                         & \perfTotalCount{}       \\
		\hline
		\multicolumn{2}{|l|}{Total}          & 665                         & \allFailureCount{}      \\
		\hline
	\end{tabular}
\end{table}

Together, the final analysis dataset contains \allFailureCount{} confirmed \skillInducedFailureClasses{}: \corrTotalCount{} \funcFails{} and \perfTotalCount{} high-confidence \effRegs{}.

\section{\funcFailMech{}}
\label{sec:rq2}

This section presents the categorization of the \corrTotalCount{} confirmed \funcFails{} in the final analysis dataset.
Table~\ref{tab:corr-roots} summarizes the functional-failure taxonomy, which consists of four high-level categories: \corrCatApplicabilityFirst{}, \corrCatEnvironmentFirst{}, \corrCatImplementationFirst{}, and \corrCatArtifactFirst{}. For categories with multiple subcategories, the table also reports their subcategory breakdowns.
Each \funcFail{} receives exactly one subcategory label, and category counts aggregate the subcategories that belong to the same mechanism family. 
For categories with a single subcategory, Table~\ref{tab:corr-roots} reports the category-level row directly.
Among the four categories, \corrCatImplementationFull{} is the largest, accounting for \corrImplementationCount{} cases (\corrImplementationPct{}), followed by \corrCatArtifactFull{} with \corrLocationCount{} cases (\corrLocationPct{}).

\begin{table*}[t]
	\caption{Classification of the \corrTotalCount{} confirmed \funcFails{}.}
	\label{tab:corr-roots}
	\centering\footnotesize
	\setlength{\tabcolsep}{3pt}
	\renewcommand{\arraystretch}{1.15}
	\begin{tabular}{|L{2.9cm}|L{3.1cm}|L{8.8cm}|r|r|}
		\hline
		\multicolumn{2}{|l|}{\textbf{Category / Subcategory}}          & \textbf{Definition}                                                                                                                                        & \textbf{Count}                                                                                                                                                                                                    & \textbf{Percentage}                              \\
		\hline
		\multicolumn{2}{|L{6.0cm}|}{\corrCatApplicabilityFull{}}                  & Skill metadata gives an incomplete or misleading applicability signal, causing the agent to apply the skill to a task where its guidance is inappropriate. & \corrMetadataCount{}                                                                                                                                                                                              & \corrMetadataPct{}                               \\
		\hline
		\multirow{3}{=}{\raggedright\arraybackslash \corrCatEnvironmentFull{}}    & \corrRootBrokenEnvFull{}                                                                                                                                   & The skill explicitly recommends or depends on a dependency or runtime component that fails in the task environment, causing installation, import, or execution failure.                                           & \corrBrokenEnvCount{}    & \corrBrokenEnvPct{}   \\
		\cline{2-5}
		                                                                           & \corrRootIncompatEnvFull{}                                                                                                                                 & The skill indirectly leads the agent to change package resolution, working directory, version, installation state, or tool state, so the agent's self-check environment state differs from the evaluator's state. & \corrIncompatEnvCount{}  & \corrIncompatEnvPct{} \\
		\cline{2-5}
		                                                                          & \multicolumn{2}{|l|}{\emph{Subtotal}}                                                                                                                      & \corrEnvironmentCount{}                                                                                                                                                                                           & \corrEnvironmentPct{}                            \\
		\hline
		\multirow{4}{=}{\raggedright\arraybackslash \corrCatImplementationFull{}} & \corrRootWorkflowFull{}                                                                                                                                    & The skill induces excessive exploration, setup, or procedural steps that prevent the agent from producing the required artifact within the available budget.                                                      & \corrWorkflowCount{}     & \corrWorkflowPct{}    \\
		\cline{2-5}
		                                                                           & \corrRootApproachFull{}                                                                                                                                    & The skill leads the agent to implement a task-required element with an incorrect method, API, value, policy, or output structure.                                                               & \corrApproachCount{}     & \corrApproachPct{}    \\
		\cline{2-5}
		                                                                           & \corrRootOmissionFull{}                                                                                                                                    & The skill leads the agent to leave a task-required element or dependent step absent, leaving the artifact incomplete without providing a concrete substitute.                                                    & \corrOmissionCount{}     & \corrOmissionPct{}    \\
		\cline{2-5}
		                                                                          & \multicolumn{2}{|l|}{\emph{Subtotal}}                                                                                                                      & \corrImplementationCount{}                                                                                                                                                                                        & \corrImplementationPct{}                         \\
		\hline
		\multicolumn{2}{|L{6.0cm}|}{\corrCatArtifactFull{}}                       & The skill leads the agent to write or integrate the required artifact at a location different from the task-specified path or integration point.           & \corrLocationCount{}                                                                                                                                                                                              & \corrLocationPct{}                               \\
		\hline
		\multicolumn{3}{|l|}{Total}                                               & \corrTotalCount{}                                                                                                                                          & \corrTotalPct{}                                                                                                                                                                                                                                                      \\
		\hline
	\end{tabular}
\end{table*}

\subsection{\corrCatApplicabilityFirst{}}

This category accounts for \corrMetadataCount{} confirmed \funcFails{} (\corrMetadataPct{}) where the skill should not have been applied under the concrete task scope.
These failures originate from the skill-loading stage: the skill's name, description, frontmatter, or when-to-use framing makes the agent treat the skill as applicable even though its guidance does not match the concrete task scope.
This category covers cases where skill metadata gives an inaccurate applicability signal.
For example, a skill written for building application-level RAG backends can make the agent treat a LangChain library-maintenance task as outside the skill's intended scope, causing it to avoid creating the task-required example files.
The decisive problem is not the final artifact path, but the decision to apply the skill under the wrong task scope.

\subsection{\corrCatEnvironmentFirst{}}

This category accounts for \corrEnvironmentCount{} confirmed \funcFails{} (\corrEnvironmentPct{}) where the skill prescribes or induces an execution environment that differs from the verifier's environment.
These failures are not primarily about what artifact should be built, but about where and under which runtime state the artifact is executed.

\noindent\emph{\corrRootBrokenEnvFirst{}} appears in \corrBrokenEnvCount{} cases (\corrBrokenEnvPct{}), where the skill explicitly recommends or depends on a package, runtime, import path, binary, or external component that fails in the task environment.

\noindent\emph{\corrRootIncompatEnvFirst{}} represents \corrIncompatEnvCount{} cases (\corrIncompatEnvPct{}) where the skill indirectly leads the agent to change package resolution, working directory, installed versions, local installs, or tool state, so the target run validates under an environment state that the evaluation environment does not share.

A representative \corrRootIncompatEnvAbbr{} case appears in an \texttt{openpyxl} task:
\begin{lstlisting}[language={}]
(*@\textbf{\texttt{Task:}}@*)
create `openpyxl/utils/report_engine.py`

(*@\textbf{\texttt{Verifier output:}}@*)
test_report_engine_importable
test_engine_generates_valid_xlsx_file
python -c "from openpyxl.utils.report_engine import *; print('OK')"

(*@\textbf{\texttt{Target trajectory:}}@*)
"Old openpyxl version incompatible ... I'll install a modern version"
(*@\textcolor{blue}{\textbf{pip install openpyxl}}@*)
(*@\textcolor{blue}{\textbf{cd /tmp}}@*)
sys.path = [p for p in sys.path if (*@\textcolor{blue}{\textbf{'/workspace/openpyxl'}}@*) not in p]

(*@\textbf{\texttt{Reference-run trajectory}:}@*)
(*@\textcolor{blue}{\textbf{patches `openpyxl/formatting/rules.py` in the repository}}@*)
`from collections import Mapping` -> `from collections.abc import Mapping`
\end{lstlisting}
The verifier imports from the repository package, while the target run validates against an external install.
This target/reference comparison shows that the failure is caused by an environment-state mismatch rather than by the artifact content alone.

\subsection{\corrCatImplementationFirst{}}

This is the largest category, accounting for \corrImplementationCount{} confirmed \funcFails{} (\corrImplementationPct{}).
These failures occur when an on-topic skill induces a fault in the implementation process or generated artifact, preventing the agent from satisfying a concrete task-required implementation element.
In many cases, the agent over-trusts a topically matched skill and treats its reusable defaults, examples, or templates as task-specific requirements.
The skill may make the workflow too heavy to finish, implement a task-required element incorrectly, or leave a required element absent.

\noindent\emph{\corrRootWorkflowFirst{}} occurs in \corrWorkflowCount{} cases (\corrWorkflowPct{}), where otherwise relevant guidance induces excessive exploration, setup, audits, or procedural checks that prevent the agent from producing the required artifact within the budget.

\noindent\emph{\corrRootApproachFirst{}} is the largest subcategory in this category, with \corrApproachCount{} cases (\corrApproachPct{}), and means that the target run implements a task-required element with an incorrect method, API, value, policy, or output structure.

\noindent\emph{\corrRootOmissionFirst{}} includes \corrOmissionCount{} cases (\corrOmissionPct{}) where the target run leaves a required field, option, behavior, domain rule, validation, or dependent step absent.
The distinction between \corrRootApproachFirst{} and \corrRootOmissionFirst{} is whether the target artifact contains a concrete but wrong implementation of the required element, or leaves that element absent.

A representative \corrRootApproachAbbr{} case appears in a spreadsheet task:
\begin{lstlisting}[language={}]
(*@\textbf{\texttt{Task:}}@*)
calculate net exports as percent of GDP

(*@\textbf{\texttt{Verifier output:}}@*)
values are about 100x too small

(*@\textbf{\texttt{Target trajectory:}}@*)
(*@\textcolor{blue}{\textbf{(Exports - Imports) / GDP}}@*)

(*@\textbf{\texttt{Reference-run trajectory:}}@*)
(*@\textcolor{blue}{\textbf{(Exports - Imports) / GDP * 100}}@*)
\end{lstlisting}
The target fills the required calculation with the wrong representation.
The reference run matters because it distinguishes a skill-induced failure from ordinary arithmetic failure: the target run uses a bare ratio, whereas the reference run keeps the task-required percentage scaling.

A representative \corrRootOmissionAbbr{} case appears in a RAG demo task:
\begin{lstlisting}[language={}]
(*@\textbf{\texttt{Task:}}@*)
chunk size, overlap, top-k, and model parameters must be configurable

(*@\textbf{\texttt{Verifier output:}}@*)
test_model_parameters_configurable

(*@\textbf{\texttt{Target trajectory:}}@*)
(*@\textcolor{blue}{\textbf{CLI args: --question, --chunk-size, --chunk-overlap, --top-k}}@*)

(*@\textbf{\texttt{Reference-run trajectory:}}@*)
(*@\textcolor{blue}{\textbf{RAGConfig includes model\_name = "gpt-3.5-turbo"}}@*)
\end{lstlisting}
The target leaves a required implementation element absent.
Here the skill example shows retrieval parameters such as chunk size, overlap, and top-k, but does not show how to configure model parameters. The target run follows that partial example: it exposes the retrieval parameters through CLI flags, but omits the required model-parameter field that the reference run adds as \texttt{model\_name}.

\finding{\textbf{Finding 1:} Only \corrMetadataCount{} of \corrTotalCount{} \funcFails{} are classified as \corrCatApplicabilityFull{} (\corrMetadataPct{}), while most fall under \corrCatImplementationFull{}, indicating that on-topic skills more often induce task-implementation faults in required implementation elements. \corrCatImplementationFull{} accounts for \corrImplementationCount{} of \corrTotalCount{} \funcFails{} (\corrImplementationPct{}), and \corrRootApproachFirst{} plus \corrRootOmissionFirst{} account for \corrApproachOmissionCount{} cases.

	\textbf{Implication:} Skill authors should separate mandatory task requirements from examples, defaults, reusable templates, and optional workflows, so that agents do not mistake reusable guidance for task-specific obligations.}

\subsection{\corrCatArtifactFirst{}}

This category accounts for \corrLocationCount{} confirmed \funcFails{} (\corrLocationPct{}) where the agent builds a plausible artifact but writes or integrates it at the wrong location.
The target run often follows repository or package-layout conventions instead of the task-specified output path.
This category covers cases where the skill leads the agent to write or integrate the required artifact at a location different from the task-specified path or integration point.

A representative case appears in a LangChain RAG task:
\begin{lstlisting}[language={}]
(*@\textbf{\texttt{Task:}}@*)
create files under (*@\textcolor{blue}{\textbf{`libs/langchain/langchain/`}}@*)
- retrievers/hybrid_retriever.py
- text_splitter/semantic_chunker.py
- chains/rag_chain.py

(*@\textbf{\texttt{Verifier output:}}@*)
file-path existence tests fail for `libs/langchain/langchain/...`

(*@\textbf{\texttt{Target trajectory:}}@*)
"The package is (*@\textcolor{blue}{\textbf{`langchain\_classic`}}@*)."
"I'll use (*@\textcolor{blue}{\textbf{`langchain\_classic`}}@*) since that's the real package"
Writes under (*@\textcolor{blue}{\textbf{`libs/langchain/langchain\_classic/`}}@*)

(*@\textbf{\texttt{Reference-run trajectory:}}@*)
"I'll create the files at the paths specified in the task"
Writes under (*@\textcolor{blue}{\textbf{`libs/langchain/langchain/`}}@*)
\end{lstlisting}
The decisive failure is artifact placement, not the plausibility of the generated RAG components.

\finding{\textbf{Finding 2:} A substantial share of \funcFails{} occur at execution-surface boundaries, where the skill changes the verifier-observed environment state or artifact location. \corrCatEnvironmentFull{} accounts for \corrEnvironmentCount{} cases (\corrEnvironmentPct{}), while \corrCatArtifactFull{} accounts for \corrLocationCount{} cases (\corrLocationPct{}).

	\textbf{Implication:} Agent platforms should treat task-specified paths, integration points, package state, and working directories as guarded constraints before accepting skill-guided environment changes or repository-native path substitutions.}

\section{\effRegMech{}}
\label{sec:rq3}

This section presents the categorization of the \perfTotalCount{} high-confidence \effRegs{} in the final analysis dataset under the primary $T=2.0$ threshold.
We use $T=2.0$ as a conservative high-confidence threshold: a run must more than double token use or execution time while the other metric also increases relative to the reference run, which separates substantial skill-induced overhead from token-time tradeoffs and ordinary run-to-run variation.
Table~\ref{tab:perf-roots} summarizes the efficiency-regression taxonomy, which consists of three high-level categories: \perfCatContextFirst{}, \perfCatProcedureFirst{}, and \perfCatDependencyFirst{}. For categories with multiple subcategories, the table also reports their subcategory breakdowns.
Each \effReg{} receives exactly one subcategory label, and category counts aggregate subcategories in the same cost-mechanism family. 
For categories with a single subcategory, Table~\ref{tab:perf-roots} reports the category-level row directly.
Among the three categories, \perfCatProcedureFull{} is the largest, accounting for \perfProcedureCount{} cases (\perfProcedurePct{}), followed by \perfCatContextFull{} with \perfContextCount{} cases (\perfContextPct{}).
The three categories correspond to three cost layers: larger per-call context, additional trajectory steps, and dependency-repair work.

\begin{table*}[t]
	\caption{Classification of the \perfTotalCount{} high-confidence \effRegs{}.}
	\label{tab:perf-roots}
	\centering\scriptsize
	\setlength{\tabcolsep}{3pt}
	\renewcommand{\arraystretch}{1.05}
	\begin{tabular}{|L{2.9cm}|L{3.1cm}|L{8.9cm}|r|r|}
		\hline
		\multicolumn{2}{|l|}{\textbf{Category / Subcategory}}     & \textbf{Definition}                                                                                                                                                   & \textbf{Count}                                                                                                                                      & \textbf{Percentage}                                 \\
		\hline
		\multirow{3}{=}{\raggedright\arraybackslash \perfCatContextFull{}}   & \perfRootBodyBloatFull{}                                                                                                                                              & The skill body adds enough text to the agent context that each model call becomes substantially more expensive.                                     & \perfBodyBloatCount{}      & \perfBodyBloatPct{}    \\
		\cline{2-5}
		                                                                     & \perfRootSuppBloatFull{}                                                                                                                                              & The skill directs the agent to load or inspect auxiliary references, templates, examples, or documentation, increasing context or tool-output cost. & \perfSuppBloatCount{}      & \perfSuppBloatPct{}    \\
		\cline{2-5}
		                                                                     & \multicolumn{2}{|l|}{\emph{Subtotal}}                                                                                                                                 & \perfContextCount{}                                                                                                                                 & \perfContextPct{}                                   \\
		\hline
		\multirow{4}{=}{\raggedright\arraybackslash \perfCatProcedureFull{}} & \perfRootExplorationFull{}                                                                                                                                            & The skill leads the agent to perform excessive repository, architecture, or pattern exploration before implementation.                              & \perfExplorationCount{}    & \perfExplorationPct{}  \\
		\cline{2-5}
		                                                                     & \perfRootPipelineFull{}                                                                                                                                               & The skill leads the agent to use a heavier construction pipeline, subprocess workflow, or runtime simulation.                      & \perfPipelineCount{}       & \perfPipelinePct{}     \\
		\cline{2-5}
		                                                                     & \perfRootVerificationFull{}                                                                                                                                           & The skill leads the agent to run excessive or repeated testing, debugging, rebuilding, or checklist verification after implementation.              & \perfVerificationCount{}   & \perfVerificationPct{} \\
		\cline{2-5}
		                                                                     & \multicolumn{2}{|l|}{\emph{Subtotal}}                                                                                                                                 & \perfProcedureCount{}                                                                                                                               & \perfProcedurePct{}                                 \\
		\hline
		\multicolumn{2}{|L{6.0cm}|}{\perfCatDependencyFull{}}                & The skill leads the agent to use a fragile or incompatible runtime dependency that requires installation, configuration, repair, or debugging before the task passes. & \perfDependencyRepairCount{}                                                                                                                        & \perfDependencyRepairPct{}                          \\
		\hline
		\multicolumn{3}{|l|}{Total}                                          & \perfTotalCount{}                                                                                                                                                     & \perfTotalPct{}                                                                                                                                                                                           \\
		\hline
	\end{tabular}
\end{table*}

\subsection{\perfCatContextFirst{}}

This category accounts for \perfContextCount{} high-confidence \effRegs{} (\perfContextPct{}) where the skill increases the cost of model calls rather than substantially changing the task-solving trajectory.
The agent may follow a similar sequence of actions as the reference run, but pays more tokens because the skill body or skill-referenced material remains in context.

\noindent\emph{\perfRootBodyBloatFirst{}} is the dominant subcategory in this category, with \perfBodyBloatCount{} cases (\perfBodyBloatPct{}), and refers to cases where the skill body itself adds enough text to make each model call substantially more expensive.

\noindent\emph{\perfRootSuppBloatFirst{}} appears in \perfSuppBloatCount{} cases (\perfSuppBloatPct{}), where the skill directs the agent to load or inspect auxiliary references, templates, examples, or documentation, expanding the context or tool-output volume.
This imbalance reflects how skills are loaded during execution: the main skill body is repeatedly present in the agent context, while supplementary material is only costly when the agent explicitly reads or follows it.
As a result, always-loaded skill text creates a broader cost surface than optional reference material.
This category suggests that skill authors should shorten mandatory skill bodies and lazy-load optional references only when the task actually requires them.

\finding{\textbf{Finding 3:} When \effRegs{} arise from context overhead, the overhead is almost entirely caused by mandatory skill-body text. \perfRootBodyBloatFirst{} accounts for \perfBodyBloatCount{} of the \perfContextCount{} context-overhead cases, while \perfRootSuppBloatFirst{} appears in only \perfSuppBloatCount{} cases.

	\textbf{Implication:} Skill packages should keep always-loaded instructions short and move examples, templates, long checklists, and background material behind explicit lazy-loading triggers.}

\subsection{\perfCatProcedureFirst{}}

This is the largest category, accounting for \perfProcedureCount{} high-confidence \effRegs{} (\perfProcedurePct{}).
These regressions occur when the skill changes the execution trajectory by adding exploration, implementation work, debugging, or verification.
These overheads often arise when skills turn optional activities into mandatory steps, including repository inspection, heavyweight construction procedures, repeated validation, and broad diagnostic checks, even when the concrete task does not require that much work.

\noindent\emph{\perfRootExplorationFirst{}} occurs in \perfExplorationCount{} cases (\perfExplorationPct{}), where the skill leads the agent to inspect architecture, search for patterns, audit related files, or compare integration points before making the main task change.

\noindent\emph{\perfRootPipelineFirst{}} represents \perfPipelineCount{} cases (\perfPipelinePct{}) where the skill induces a heavier construction process, such as multi-stage conversion, subprocess workflows, or runtime simulation.

\noindent\emph{\perfRootVerificationFirst{}} is the largest subcategory, with \perfVerificationCount{} cases (\perfVerificationPct{}), and refers to excessive or repeated tests, rebuilds, debugging attempts, or checklist verification after the main artifact has been produced.

\finding{\textbf{Finding 4:} High-confidence \effRegs{} are dominated by \perfCatProcedureFull{} rather than prompt length alone, and the largest sources within \perfCatProcedureFull{} are excessive verification and heavy implementation pipelines. At $T=2.0$, \perfCatProcedureFull{} accounts for \perfProcedureCount{} of \perfTotalCount{} \effRegs{} (\perfProcedurePct{}); within this category, \perfRootVerificationFirst{} contributes \perfVerificationCount{} cases and \perfRootPipelineFirst{} contributes \perfPipelineCount{} cases.
	
	\textbf{Implication:} Skill platforms should model the extra actions that skills induce, and skill authors should condition verification scope and pipeline depth on task uncertainty, change size, and budget rather than prescribing exhaustive workflows by default.}

We distinguish \perfCatContextFull{} from \perfCatProcedureFull{} by asking where the extra cost comes from.
A case is \perfCatContextFull{} when each model call becomes more expensive because the skill body or supplementary material increases context size, while the overall sequence of task-solving steps remains similar to the reference run.
A case is \perfCatProcedureFull{} when the skill changes the trajectory by adding extra exploration, implementation work, retries, or verification steps.
If both effects appear, we assign the subcategory according to the larger cost driver observed in the trajectory.

\subsection{\perfCatDependencyFirst{}}

This category accounts for \perfDependencyRepairCount{} high-confidence \effRegs{} (\perfDependencyRepairPct{}) where the task eventually passes, but the skill leads the agent to use a fragile or incompatible runtime dependency that requires installation, configuration, repair, or debugging.
This category covers cases where dependency setup or repair becomes part of the successful trajectory.
If the same dependency problem remains unresolved and causes the verifier to fail, it appears in the \funcFailTax{} as \corrRootBrokenEnvFirst{}.

\section{Automated Attribution of \SkillInducedFailures{}}
\label{sec:attribution}

As reusable skills become part of agent platforms and skill marketplaces, failure analysis must eventually move beyond one-off manual audits.
To explore this use case, we build \textsc{SkillTriage}, a lightweight LLM-based tool for post-confirmation triage: given a target/reference pair already identified as a \skillInducedFailure{}, it predicts the high-level category and subcategory, and returns contrastive evidence for that attribution.
Figure~\ref{fig:attribution-overview} gives an overview of its input, differential-evidence, and attribution stages.
We evaluate \textsc{SkillTriage} against manually assigned labels for both \funcFails{} and \effRegs{}; the tool is not used to create the taxonomy or replace manual validation.

\subsection{Attribution Tool}

The key insight behind \textsc{SkillTriage} is that the manual taxonomy can be operationalized as an attribution procedure, not merely used as a set of labels.
The diagnostic principle is to ask which label best explains both the target outcome, either failure or extra cost, and the divergence between the target-run and reference-run trajectories.
Thus each category acts as an evidence checklist: \funcFails{} require skill-scope, runtime, artifact-construction, or task-path evidence, while \effRegs{} require context, procedure, or dependency-cost evidence.

\begin{figure}[t]
\centering
\includegraphics[width=\linewidth]{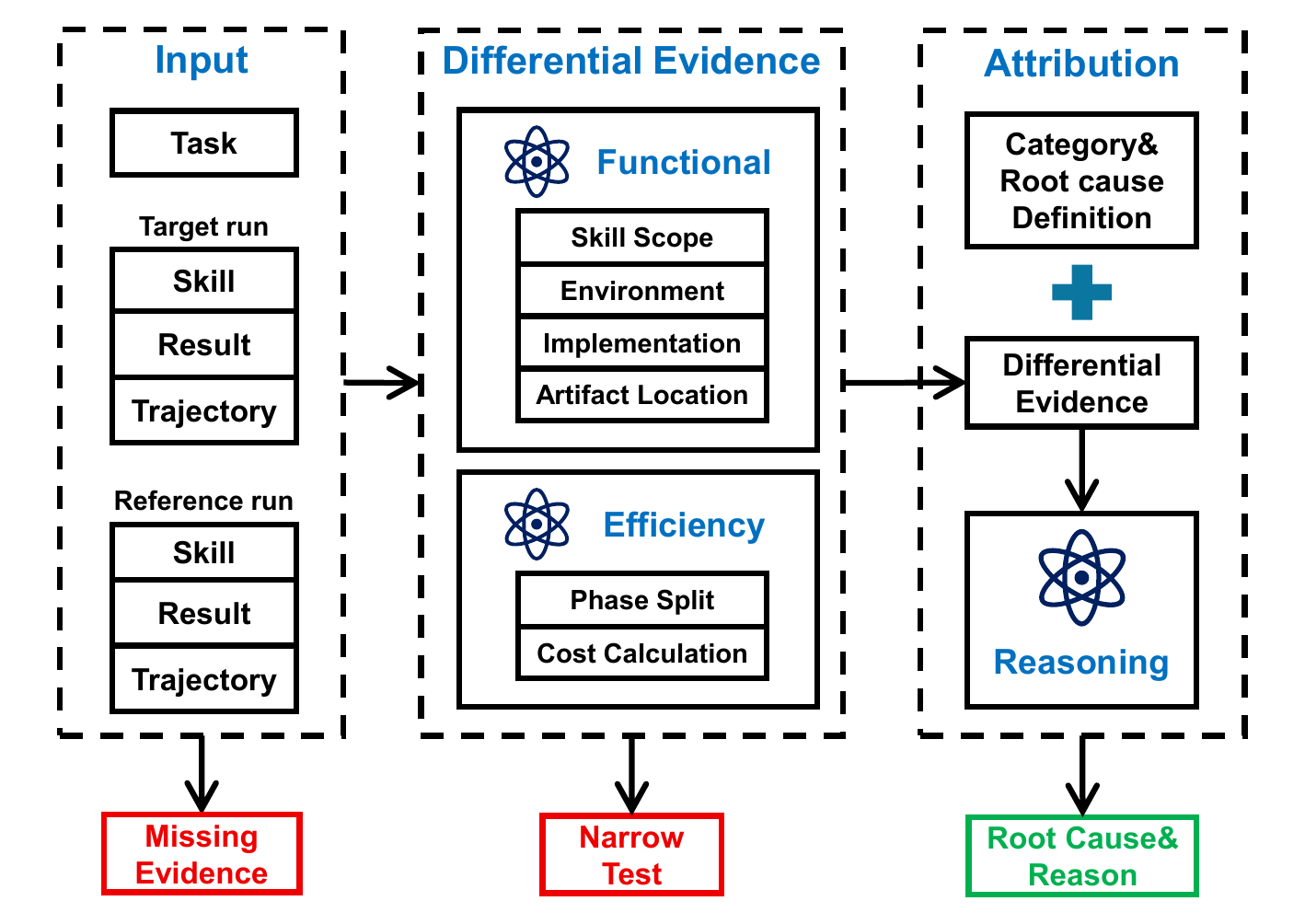}
\caption{Workflow of \textsc{SkillTriage}.}
\label{fig:attribution-overview}
\end{figure}

Figure~\ref{fig:attribution-overview} organizes this taxonomy-guided procedure into three stages: input construction, differential-evidence extraction, and attribution.

\noindent\textit{Input construction.}
\textsc{SkillTriage} first normalizes each paired case into a shared task view and two run views: the target-run view records the loaded skill, result, and trajectory, while the reference-run view records the no-skill or matched-skill setup, result, and trajectory. At this stage, deterministic evidence gates check whether the case has the minimum trajectory, skill, and verifier evidence needed for attribution.

\noindent\textit{Differential evidence.}
\textsc{SkillTriage} then turns taxonomy boundaries into explicit differential signals rather than asking the model to infer all evidence directly from raw traces.
For \funcFails{}, \textsc{SkillTriage} computes five differential signals (DS1--DS5) over the functional-evidence surfaces in Figure~\ref{fig:attribution-overview}: skill scope, environment, implementation, and artifact location. DS1 tests whether target-only environment or runtime-state changes plausibly explain the verifier failure. DS2 tests whether such changes are explicitly skill-prescribed or indirectly skill-induced. DS3 tests task-required paths against target write paths. DS4 tests whether the target run produced, stalled before producing, or refused to produce the required artifact. DS5 uses a focused construction-difference check to separate incorrect implementation of a task-required element from required-element omission.
For \effRegs{}, \textsc{SkillTriage} computes phase-level and action-tag cost evidence from step-level token/time/tool records. 
It splits steps into broad phases such as pre-implementation exploration, implementation or answer-producing pipeline work, and post-implementation verification/debugging, with dependency/setup steps tracked when install, import, or environment-repair signals appear.
It also assigns action tags such as exploration (read/search), pipeline (write/data-processing commands), verification (test/debug/rebuild/check), dependency repair (install/import/environment fixes), and reference loading (reading skill-linked templates or documents). 
The tool then computes aggregate, phase-level, and tag-level cost deltas and uses these signals to support context, procedure, and dependency-cost evidence. 
The preliminary cost-dominance estimate is used only as an advisory hint; the final label is selected by comparing the high-cost and low-cost step-level evidence and identifying the mechanism that best explains the extra token/time cost.

\noindent\textit{Attribution.}
The attribution stage gives the model the taxonomy definitions, normalized input, and extracted differential evidence. It then reasons over candidate labels and selects the root cause that best explains both the target outcome and the target/reference trajectory divergence.
For example, when DS3 shows that the target run writes a plausible artifact to a path different from the task-specified path while the reference run writes to the expected path, the attribution stage selects \corrRootLocationFull{} rather than implementation-content subcategories because the target-run/reference-run divergence is artifact placement, not artifact content.
In both branches, the report returns a category, subcategory, natural-language reason, cited skill section or trajectory evidence, and a repair suggestion, so the output supports debugging rather than only label prediction.

\subsection{Evaluation Setup}
We evaluate \textsc{SkillTriage} separately on functional failures and efficiency regressions.
For \funcFails{}, we run the tool on all \corrTotalCount{} confirmed cases validated by the manual audit; this set includes both with/no-skill and cross-skill failures and covers all seven functional-failure subcategories.
For \effRegs{}, we run the tool on all \perfTotalCount{} high-confidence \textsc{PASS}/\textsc{PASS} regressions at the primary $T=2.0$ threshold, including \perfBaselineCount{} with/no-skill cases and \perfCrossSkillCount{} cross-skill cases.
For both case sets, we do not rerun the benchmark tasks; instead, we run the attribution procedure with GPT-5.5~\cite{openai2026gpt55} three times on the same paired execution evidence for each case and aggregate the three outputs with a 2-of-3 majority vote.
Both evaluations use manually assigned attribution labels derived from the corresponding functional-failure and efficiency-regression audits.
Majority voting is a standard ensemble strategy for reducing prediction variance~\cite{dietterich2000ensemble}; in LLM settings, related self-consistency methods similarly aggregate multiple sampled outputs to improve reliability~\cite{wang2023selfconsistency}.
We use a 2-of-3 vote because three runs are the smallest odd-number setting that supports a majority decision, reducing single-run attribution variance while keeping evaluation cost modest.

We report three metrics for each branch.
Exact subcategory accuracy measures whether the predicted subcategory matches the manual label.
Category accuracy measures whether the prediction falls into the correct high-level category.
For \funcFails{}, we also report exact subcategory accuracy within \corrCatImplementationFirst{} cases, the largest functional-failure category.
For \effRegs{}, we analogously report exact subcategory accuracy within \perfCatProcedureFirst{} cases, the largest performance category and a challenging attribution subset, because it requires distinguishing exploration, implementation-pipeline, and verification overhead.

\subsection{Attribution Results}

Overall, \textsc{SkillTriage} provides useful triage-level attribution: it recovers high-level categories for most \skillInducedFailures{} and identifies exact subcategories in a substantial majority of cases, while the remaining errors concentrate near taxonomy boundaries.
Table~\ref{tab:attribution-accuracy} reports 2-of-3 majority agreement for \funcFail{} attribution over three independent attribution runs.
\textsc{SkillTriage} matches the human-audit high-level category for \attribVoteCombinedCategory{} cases (\attribVoteCombinedCategoryPct{}) and the exact subcategory for \attribVoteCombinedRoot{} cases (\attribVoteCombinedRootPct{}).
On \corrCatImplementationAbbr{} cases, where it must distinguish \corrRootWorkflowFirst{}, \corrRootApproachFirst{}, and \corrRootOmissionFirst{}, it reaches \attribVoteCombinedCthree{} (\attribVoteCombinedCthreePct{}).
The per-root recall inside \corrCatImplementationAbbr{} is 3/4 for \corrRootWorkflowAbbr{}, 41/46 for \corrRootApproachAbbr{}, and 32/36 for \corrRootOmissionAbbr{}.

\begin{table}[t]
\caption{\funcFail{} attribution results.}
\label{tab:attribution-accuracy}
\centering\footnotesize
\setlength{\tabcolsep}{3pt}
\renewcommand{\arraystretch}{1.12}
\begin{tabular}{|L{1.5cm}|R{2.3cm}|R{2.3cm}|R{1.8cm}|}
\hline
\textbf{Subset} & \textbf{Subcategory} & \textbf{Category} & \textbf{\corrCatImplementationAbbr{} subcat.} \\
\hline
With/no-skill & \attribVoteBaselineRoot{} (\attribVoteBaselineRootPct{}) & \attribVoteBaselineCategory{} (\attribVoteBaselineCategoryPct{}) & \attribVoteBaselineCthree{} (\attribVoteBaselineCthreePct{}) \\
\hline
Cross-skill & \attribVoteCrossRoot{} (\attribVoteCrossRootPct{}) & \attribVoteCrossCategory{} (\attribVoteCrossCategoryPct{}) & \attribVoteCrossCthree{} (\attribVoteCrossCthreePct{}) \\
\hline
Combined & \attribVoteCombinedRoot{} (\attribVoteCombinedRootPct{}) & \attribVoteCombinedCategory{} (\attribVoteCombinedCategoryPct{}) & \attribVoteCombinedCthree{} (\attribVoteCombinedCthreePct{}) \\
\hline
\end{tabular}
\end{table}

Table~\ref{tab:performance-attribution-accuracy} reports 2-of-3 majority-vote agreement for \effReg{} attribution against the manual performance labels.
Across \perfTotalCount{} high-confidence \effRegs{}, \textsc{SkillTriage} matches the manual performance category for 145/182 cases (79.7\%) and the exact subcategory for 132/182 cases (72.5\%).
Within the \perfCatProcedureAbbr{} subset, the tool reaches 78/114 exact subcategory agreement (68.4\%), reflecting the difficulty of separating \perfRootExplorationFirst{}, \perfRootPipelineFirst{}, and \perfRootVerificationFirst{}.

\begin{table}[t]
\caption{\effReg{} attribution results.}
\label{tab:performance-attribution-accuracy}
\centering\footnotesize
\setlength{\tabcolsep}{3pt}
\renewcommand{\arraystretch}{1.12}
\begin{tabular}{|L{1.5cm}|R{2.3cm}|R{2.3cm}|R{1.8cm}|}
\hline
\textbf{Subset} & \textbf{Subcategory} & \textbf{Category} & \textbf{\perfCatProcedureAbbr{} subcat.} \\
\hline
With/no-skill & 98/128 (76.6\%) & 102/128 (79.7\%) & 49/69 (71.0\%) \\
\hline
Cross-skill & 34/54 (63.0\%) & 43/54 (79.6\%) & 29/45 (64.4\%) \\
\hline
Combined & 132/182 (72.5\%) & 145/182 (79.7\%) & 78/114 (68.4\%) \\
\hline
\end{tabular}
\end{table}

The residual errors in both branches are boundary errors: the same differential evidence can support neighboring taxonomy labels.
For \funcFails{}, the 2-of-3 majority leaves 14/125 exact-subcategory errors, concentrated near the \corrRootApproachAbbr{}/\corrRootOmissionAbbr{} and \corrCatEnvironmentFirst{}/\corrRootLocationFirst{} taxonomy boundaries; for example, when a required helper function is implicit in the task, the tool may classify its absence as an incorrect implementation of nearby code rather than as a required-element omission.
For \effRegs{}, the 50/182 exact-subcategory errors often arise when one high-cost trajectory contains several plausible cost surfaces: dependency repair inside a verification loop, skill-body cost inflating test steps, or build commands that could be implementation or verification.
This error structure suggests that automated attribution should expose the exact skill section, trajectory step, artifact difference, or cost-heavy step used as evidence, not only the final label.

\section{Discussion}
\label{sec:discussion}

\subsection{Future Research Directions}
\label{sec:research}
Our findings suggest three directions for making skill reuse safer and more cost-aware.

\noindent\textit{Skill-task compatibility checks.}
Because many \funcFails{} arise from topically relevant skills that distort task-required implementation elements or verifier-visible surfaces, future platforms should compare task requirements with candidate skill contents before and during execution.
Such checks can extract required fields, APIs, paths, output formats, domain rules, and environment constraints, then warn when a skill introduces conflicting defaults, examples, templates, paths, or package-state assumptions.

\noindent\textit{Cost-aware skill packaging and selection.}
Our \effReg{} findings show that skill cost comes from both always-loaded context and induced trajectory steps, not prompt length alone.
Future tools should estimate marginal context cost, move task-specific examples or checklists to lazy-loaded references, and predict extra exploration, implementation, verification, or dependency-repair work before selecting a skill.

\noindent\textit{Budget-aware execution policies.}
The largest sources within \perfCatProcedureFull{} are excessive verification and heavy implementation pipelines, suggesting that agents need explicit policies for adapting verification scope and implementation-pipeline depth to task uncertainty, repository size, change risk, and remaining token or execution-time budget.
Such policies would preserve careful skill-guided workflows while preventing optional guidance from becoming mandatory work on every task.

\subsection{Threats to Validity}
\label{sec:threats}

\emph{Threats to Internal Validity}.
Despite the availability of reference materials, such as task specifications, verifier outputs, skill files, generated artifacts, and agent trajectories, analyzing \funcFails{}, \effRegs{}, and their root causes requires manual judgment.
The complexity of agent trajectories introduces subjectivity into deciding whether a verifier is narrower than the task, whether a cost change reflects ordinary agent variance, and how the target run diverges from the reference run.
To reduce this threat, we excluded ambiguous and verifier-narrow cases and reached group consensus before finalizing the taxonomy labels.
Moreover, \textsc{SkillTriage} provides an independent consistency check by comparing taxonomy-guided triage reports with manually assigned attribution labels for both \funcFails{} and \effRegs{}.

\emph{Threats to External Validity}.
Our study covers agent-skill tasks from multiple application domains, including software engineering, healthcare, manufacturing, cybersecurity, natural science, finance, and mathematics.
The dataset includes verifier-graded tasks from SkillsBench and repository-based tasks with deterministic tests from SWE-Skills-Bench, allowing us to analyze skill behavior through reproducible pass/fail and cost signals.
This setting provides broader coverage than a single application domain and includes both general task-solving and software-engineering scenarios.
However, it is possible that certain findings depend on the task distribution, agent harness, model, or skill ecosystem used in our study and may not directly apply to all settings.
To mitigate this threat, we sought to avoid drawing conclusions that are specific to these two benchmarks.

% related_work.tex
% Related Work section for the main paper.
% Requires a bibliography file (related_work.bib) to be loaded by the main paper.

\section{Related Work}
\label{sec:related}

The \emph{Agent Skills} abstraction~\cite{anthropic2025agentskills} packages reusable procedural knowledge as on-demand \texttt{SKILL.md} documents, extending work on agent capabilities such as skill libraries, executable actions, and tool-oriented tuning~\cite{wang2023voyager,wang2024codeact,yang2023gpt4tools}. SkillsBench~\cite{li2026skillsbench} and SWE-Skills-Bench~\cite{han2026sweskillsbench} show that skills can improve average pass rate but also introduce regressions, zero-gain cases and token overhead. We build on these benchmarks but study a different question: which skill contents and induced trajectory changes cause \funcFails{} and \effRegs{}.

Skill loading is also a form of context engineering inside the agent reasoning-and-acting loop~\cite{yao2023react}. Prior work shows that additional context is not monotonically helpful: models can ignore middle context, become distracted by irrelevant content, degrade with more documents or longer inputs, and react strongly to prompt format~\cite{liu2024lostmiddle,shi2023distracted,levy2025moredocs,levy2024sametask,sclar2024sensitivity}. These studies motivate skill-loading risk, while our work analyzes how loaded skill context changes executable repository actions, implementation choices, verification behavior, and cost.

Empirical agent studies increasingly analyze failures from traces and benchmarks. \textsc{MAST}~\cite{cemri2025mast} builds a taxonomy from agent traces; Agentless~\cite{xia2024agentless} and SWE-agent~\cite{yang2024sweagent} show that scaffold design can materially change outcomes; SWE-bench and its extensions provide evaluation infrastructure for software agents~\cite{jimenez2024swebench,wang2024openhands,zan2025multiswebench,miserendino2025swelancer,kwa2025metr}. Our attribution target is different: we attribute failures and cost regressions to loaded skills under controlled contrastive comparisons, rather than to the base agent or task alone.

Our taxonomy also draws on empirical software-engineering studies of performance bugs, ML bugs, and configuration bugs~\cite{jin2012perf,nistor2013perf,selakovic2016jsperf,cao2022dlperf,islam2019dlbugs,humbatova2020dltax,zhang2018tfbugs,yin2011confbugs,xu2015knobs,han2016confperf}. We adapt these perspectives to a new unit of analysis: agent trajectories shaped by skill.

\section{Conclusion}
\label{sec:concl}
\enlargethispage{\baselineskip}

This paper studied \emph{\skillInducedFailures{}}, a failure class induced by skills in LLM agents. 
By constructing a contrastive analysis dataset, we identified \allFailureCount{} confirmed cases: \corrTotalCount{} \funcFails{} and \perfTotalCount{} high-confidence \effRegs{}. 
Our analysis shows that skills induce \funcFails{} through applicability, environment, task-implementation, and artifact mechanisms, and induce \effRegs{} mainly through \perfCatProcedureFull{} rather than prompt length alone. 
These findings suggest that skill loading should be treated as a costed and validated configuration decision, supported by compatibility checks, cost prediction, and paired correctness/cost evaluation.
Moreover, we developed \textsc{SkillTriage}, a taxonomy-guided triage tool that normalizes paired cases, extracts differential evidence, and attributes confirmed failures and regressions to root causes.

\clearpage

% \ifdefined\includebibliography
\bibliographystyle{IEEEtran}
\bibliography{related_work}
% \fi

\end{document}